\pdfoutput=1

\documentclass[11pt]{article}

\usepackage[final]{acl}

\usepackage{times}
\usepackage{latexsym}

\usepackage[T1]{fontenc}

\usepackage[utf8]{inputenc}

\usepackage{microtype}

\usepackage{inconsolata}

\usepackage{array}
\usepackage{multirow}
\usepackage{xspace}
\usepackage{enumerate}
\usepackage{pifont}
\usepackage{paralist}
\usepackage{amsmath}
\usepackage{graphicx}
\usepackage{subcaption}
\usepackage{xcolor}
\usepackage{colortbl}
\usepackage[normalem]{ulem}

\definecolor{gray}{gray}{0.9}

\usepackage{booktabs}

\title{EFUF: Efficient Fine-Grained Unlearning Framework for Mitigating Hallucinations in Multimodal Large Language Models}

\author{
    Shangyu Xing \quad Fei Zhao \quad Zhen Wu{\thanks{~~~Corresponding author.}} \quad Tuo An \\
    {\bf Weihao Chen} \quad {\bf Chunhui Li} \quad {\bf Jianbing Zhang} \quad {\bf Xinyu Dai} \\
    National Key Laboratory for Novel Software Technology, Nanjing University, China \\
    {\tt \{xsy, zhaof, ant, chenwh, lich\}@smail.nju.edu.cn} \\
    {\tt \{wuz, zjb, daixinyu\}@nju.edu.cn} \\
}

\begin{document}
\maketitle
\begin{abstract}

Multimodal large language models (MLLMs) have attracted increasing attention in the past few years, but they may still generate descriptions that include objects not present in the corresponding images, a phenomenon known as object hallucination. To eliminate hallucinations, existing methods manually annotate paired responses with and without hallucinations, and then employ various alignment algorithms to improve the alignment capability between images and text. However, they not only demand considerable computation resources during the finetuning stage but also require expensive human annotation to construct paired data needed by the alignment algorithms. To address these issues, we propose an efficient fine-grained unlearning framework (EFUF), which performs gradient ascent utilizing three tailored losses  to eliminate hallucinations without paired data. Extensive experiments show that our method consistently reduces hallucinations while preserving the generation quality with modest computational overhead. Our code and datasets are available at \href{https://github.com/starreeze/efuf}{https://github.com/starreeze/efuf}.






\end{abstract}

\section{Introduction}
\begin{figure}[t]
    \centering
    \includegraphics[width=0.5\textwidth]{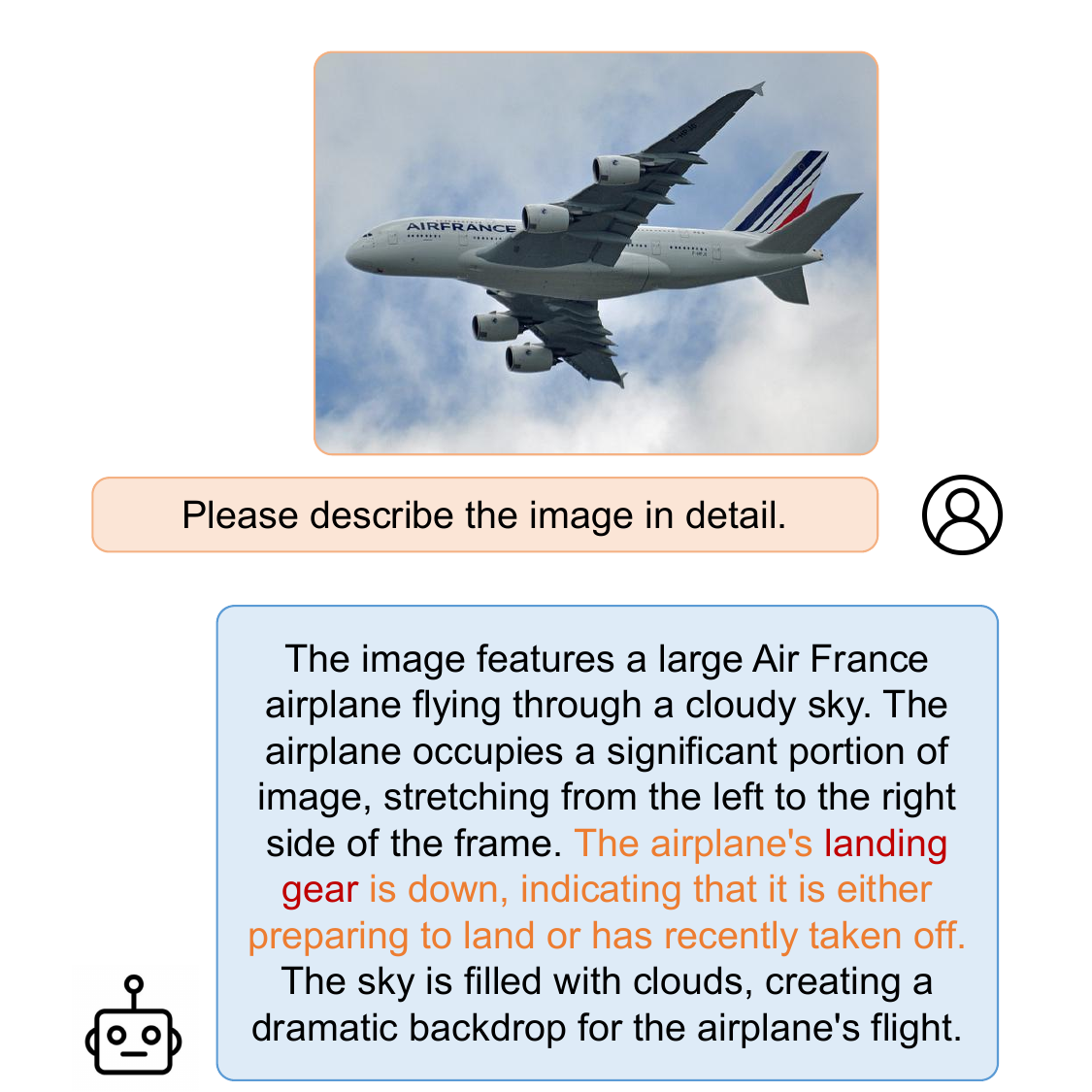}
    \caption{An example of hallucination in MLLM.}
    \label{fig:example}
\end{figure}

In the burgeoning field of artificial intelligence, the advent of multimodal large language models (MLLMs) has opened new frontiers in human-computer interaction, data processing, and automated content generation \citep{minigpt, llava, share4v, owl}. These sophisticated models, capable of understanding both text and images, have significantly advanced our ability to automate complex tasks.

However, an intriguing and critical phenomenon known as ``hallucination'' in these models poses unique challenges for current research. Hallucination in MLLMs refers to the generation of inconsistent responses that are not grounded by the multimodal context \citep{1}. For example, as shown in Figure \ref{fig:example}, the caption includes the object ``landing gear'', but in fact it does not appear in the image. Such hallucinations will lead to misinformation, potentially undermining user trust in numerous downstream applications.


Recent methods for mitigating multimodal hallucination can be divided into two categories: inference-based methods \citep{6,7,8,9,vcd,icd,hio} and finetuning-based methods \citep{1,2,3,4,5}. Inference-based methods correct or restrict generated content through external expert review, self-reflection or decoding strategies during inference stage. However, they usually require additional inference steps with increased costs and delay \citep{2}. Furthermore, each task demands specific procedure or prompt \citep{10}, adding to the complexity of implementation. Overcoming these drawbacks, finetuning-based approaches are proposed to adjust the model directly through specialized datasets and preference alignment algorithms. These algorithms, including RLHF \citep{1,3}, DPO \citep{2,4,zhou2024aligning} and contrastive learning \citep{5}, enhance the congruence between text and images, leading to improved alignment.
Although they have achieved good performance, two critical issues emerge:

First, their data demands are substantial, as they require a comprehensive set of paired positive and negative samples for effective finetuning. The alignment algorithms they employed demand paired hallucinated and non-hallucinated responses for each query. Acquiring such specific and varied response sets for each query presents a significant challenge. Recent methodologies in this field predominantly rely on human labor to annotate the output from the MLLM, requiring specialized expertise and incurring considerable expenditure of time and financial resources.

Second, The finetuning of MLLM utilizing these alignment algorithms usually demands considerable computational resources. Most of these techniques are sophisticated and necessitate the simultaneous operation of multiple models to execute preference alignment, thereby escalating the overall cost significantly.

To tackle the above issues, we 
propose the \textbf{E}fficient \textbf{F}ine-Grained \textbf{U}nlearning \textbf{F}ramework (EFUF), which offers the advantage of not necessitating paired data and being more efficient during the finetuning phase. Our method, grounded in the principles of unlearning, mainly relies on performing gradient ascent on negative samples to mitigate hallucinations, eliminating the need for costly manually-annotated paired data. Additionally, 
it consumes considerably fewer computational resources. Unlike traditional alignment algorithms that require simultaneous operation of multiple models to execute preference alignment, EFUF operates without this requirement.

The key to applying the unlearning algorithm is how to curate positive and negative samples, i.e., distinguish between real and hallucinated objects, in a manner that is both cost-effective and reliable. Intuitively, the similarity between objects and their corresponding images can act as an indicator for hallucinations, since the image contains real objects but not the hallucinated ones. Inspired by \citet{zhao2024aligngpt}, we propose to utilize the CLIP model \citep{DBLP:conf/icml/RadfordKHRGASAM21} to evaluate text-image congruence. Trained on a vast corpus of text-image pairs, CLIP stands as a robust tool to help identify hallucinations.



After ascertaining the capability of CLIP through a preliminary experiment, we curate our dataset manually-free by utilizing CLIP scores, before applying our unlearning-based method to MLLMs. This process enables us to harness the power of unlearning, offering a potent and efficient approach for mitigating hallucinations in MLLMs.

Our contribution can be summarized as follows:


\begin{compactenum}[1)]
    \item To the best of our knowledge, we provide a new perspective to utilize unlearning to mitigate multimodal hallucination in MLLMs.
    
    \item We propose an efficient fine-grained unlearning framework EFUF, which can obtain positive and negative examples separately in a cost-effective and reliable manner.
    
    \item EFUF has good compatibility and can be easily extended to existing MLLMs. Experiments conducted across a range of MLLMs validate the effectiveness of our method.
\end{compactenum}


\section{Related Work}
In this section, we review the existing studies on Hallucination Mitigation for MLLM and Unlearning algorithm.

\subsection{Hallucination Mitigation for MLLM}
To mitigate hallucinations for MLLM, various methods have been proposed. According to different phase during which they tackle the hallucinations, their work can be divided into two categories:

(1) Inference-based methods. They employ external experts, self-reflection framework or decoding strategies to constrain or modify generated content during the inference phase, thereby reducing hallucinations. For example, LURE \citep{7} utilizes manually-crafted features to detect hallucinations and therefore revises the generated text. Woodpecker \citep{8} proposes to post-edit hallucinations by combining the output of MLLMs and a more accurate expert VQA model using GPT-3.5. VIGC \citep{9} iteratively refines the instruction data using generation and correction framework. VOLCANO \citep{6} trains the MLLM to give self-feedback, and then performs self-reflection on the original generated text according to the feedback. VCD \citep{vcd} first introduces contrastive decoding in MLLM by disturbing the visual inputs and calculate visual uncertainty to restrict the generation of hallucinated tokens. ICD \citep{icd} utilizes disturbance on instructions instead of images. HIO \citep{hio} employs a hallucinated model to further widen the gap between hallucinated and correct tokens, achieving better contrastive outcomes. Although these methods do not need to train the model, they require additional inference steps with increased costs and delay \citep{2}, and specific procedure and prompt must be designed for each task \citep{10}.

(2) Finetuning-based methods. Overcoming the potential drawbacks of the first category, these methods involve crafting specific datasets and finetuning the model, aiming for better alignment between images and text. For instance, LLaVA-RLHF \citep{1} first adopts RLHF to mitigate hallucinations. Based on this work, RLHF-V \citep{2} introduces fine-grained alignment by manually correcting the outputs of MLLMs. Beyond standard RLHF, some works utilize other improved algorithms for better efficiency, e.g., DPO \citep{4, zhou2024aligning}, instruction tuning \citep{3}, and contrastive learning \citep{5}. However, these methods require expensive manually annotated paired data, and most of them also demand substantial computational resources during the finetuning stage. Therefore, in this work, we focus on reducing the data and computation requirements.

\subsection{Unlearning}
Unlearning refers to a technique designed to induce a model to "forget" specific behaviors or data, primarily through the application of gradient ascent methods \citep{7163042}. Recently, unlearning for LLM is receiving increasing attention. \citet{jang-etal-2023-knowledge} demonstrate that straightforward gradient ascent can effectively eliminate privacy vulnerabilities in LLMs. Later, \citet{yao2023large} propose the use of random mismatch and restrictions on KL divergence for positive samples, reducing the negative impact of unlearning on the general performance of LLMs.

In our research, we extend the concept of unlearning to the realm of multimodal hallucination mitigation in MLLMs, proposing a novel solution for enhancing model reliability and accuracy in multimodal contexts. 
In contrast to earlier approaches that apply unlearning across the entirety of a model's responses, our methodology focuses exclusively on the unlearning of hallucinated objects. This precise, fine-grained unlearning strategy allows for a more sophisticated refinement of the model's outputs, ensuring that only inaccuracies are corrected without diminishing the model's capabilities in other areas. To the best of our knowledge, this is the first attempt to adopt unlearning to multimodal large language models.

\section{Preliminary Experiment}\label{Preliminary_Experiment}
\begin{figure*}[t]
    \centering
    \begin{subfigure}{0.45\textwidth}
        \includegraphics[width=0.9\textwidth]{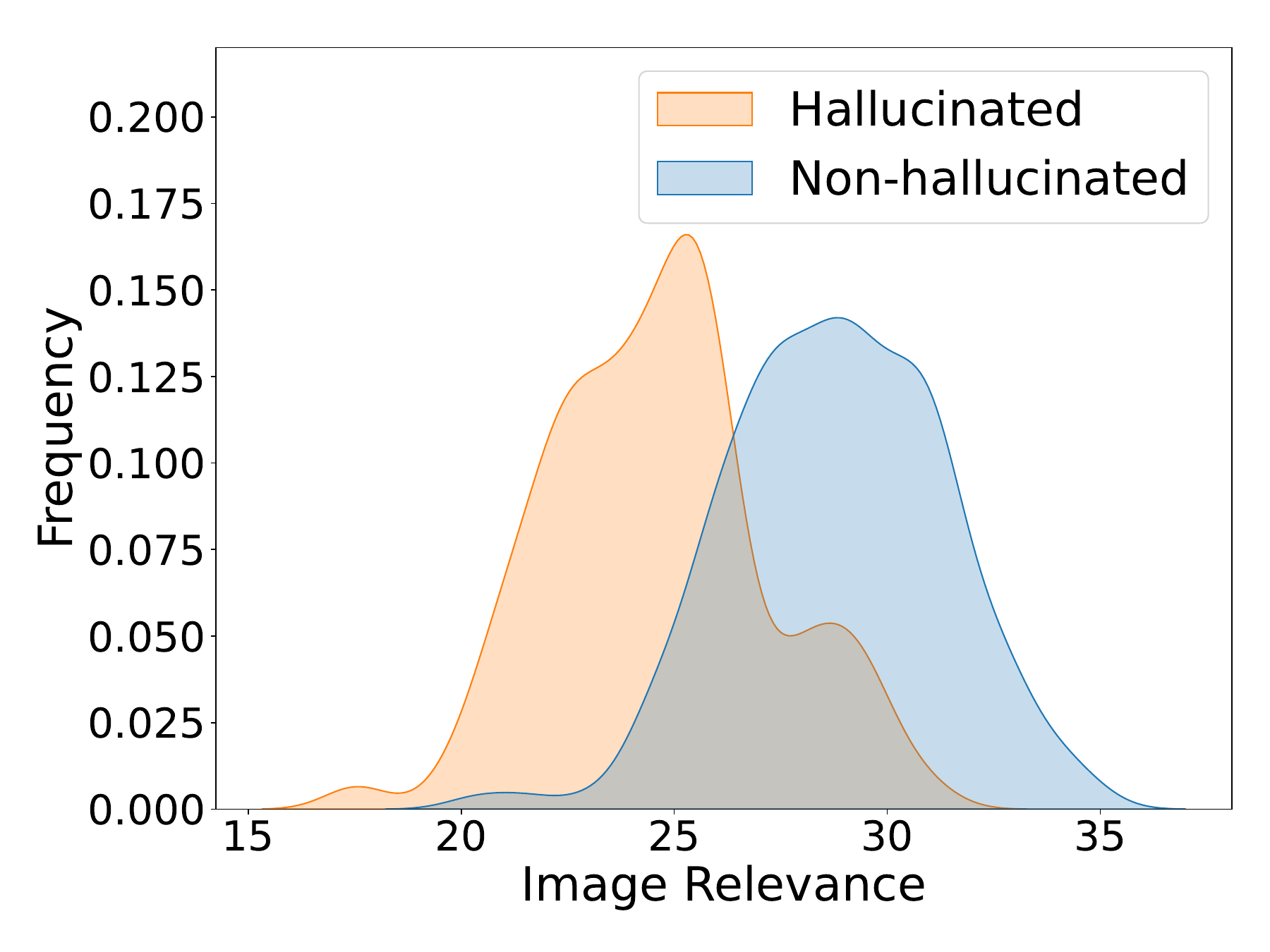}
        \caption{MiniGPT4}
    \end{subfigure}
    \begin{subfigure}{0.45\textwidth}
        \includegraphics[width=0.9\textwidth]{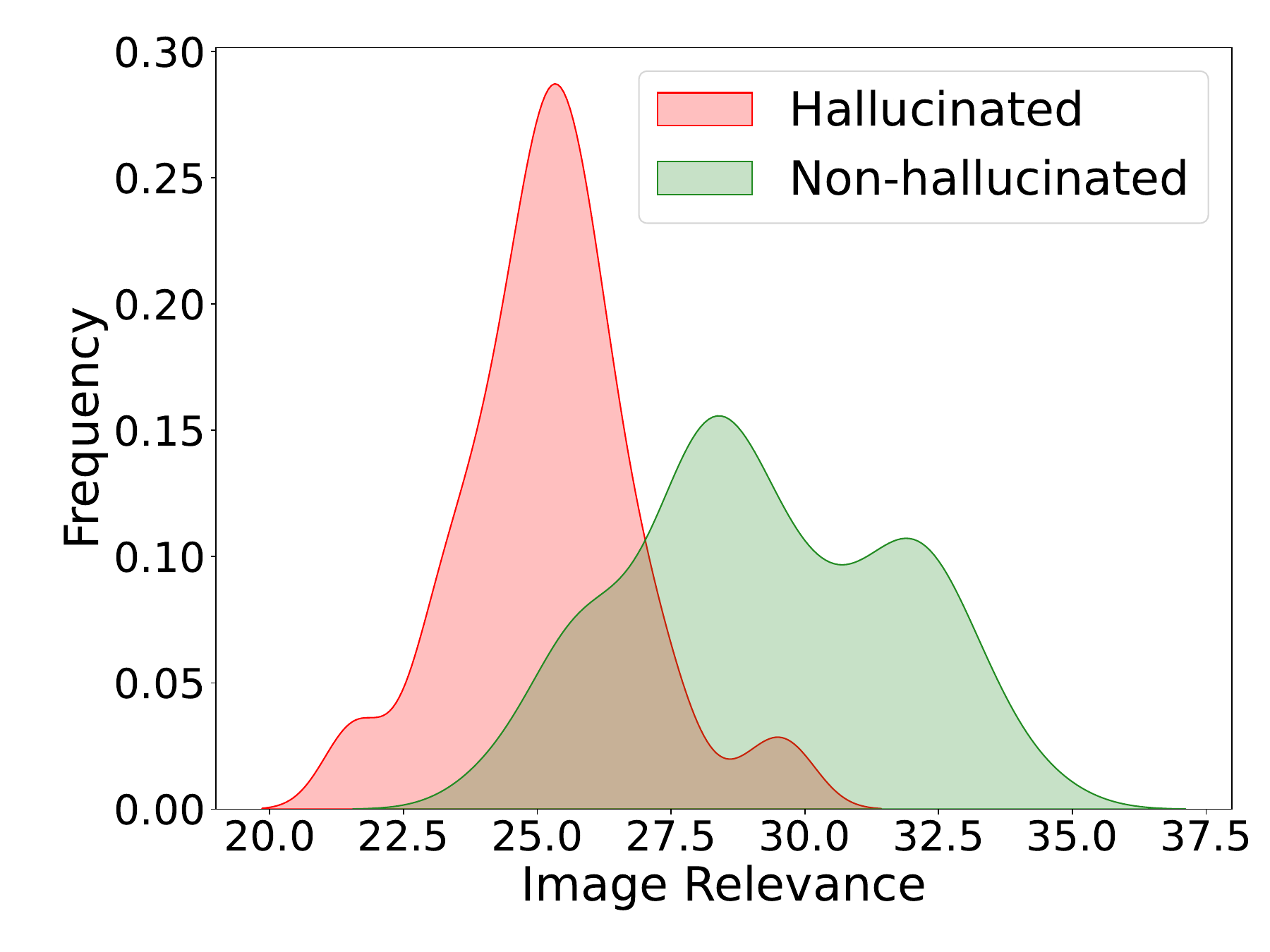}
        \caption{LLaVA}
    \end{subfigure}
    \caption{Comparison of hallucinated and non-hallucinated objects generated by MiniGPT4 (a) and LLaVA (b) on image-relevance scores.}
    \label{fig:pre}
\end{figure*}

The initial phase of our research involves confirming the hypothesis that text-image congruence can serve as a reliable indicator of hallucination occurrences. To this end, we designed a preliminary study aimed at validating this premise. Below, we detail the methods and findings of this experiment.

\subsection{Hallucinated \textit{v.s.} Non-Hallucinated}
Our approach involves employing the CLIP model to assess the similarity between text and corresponding images, with the objective of determining whether there is a discernible difference in the similarity scores of hallucinated versus non-hallucinated content. Following \citet{7}, we manually annotate 200 image captions generated by MiniGPT \citep{minigpt} and LLaVA \citep{llava}, labeling objects as either hallucinated or non-hallucinated. Subsequently, we define an object-level image-relevance score by calculating fine-grained CLIP similarities for these objects in relation to their associated image segments, aiming to uncover any significant disparities in score distributions.

Formally, let $V = \{v_1, v_2, ..., v_m\}$ denotes the collection of images, and $T = \{t_1, t_2, ..., t_m\}$ is the corresponding captions generated by the MLLM. For each $t_i \in T$, we manually annotated all the objects in the caption, represented by $O_i = \{o_i^1, o_i^2, ..., o_i^n\}$, and $O = \{O_1, O_2, ..., O_m\}$. After that, we determine whether the object is hallucinated, i.e., whether it appears in the image, assigning each object a binary value $h(o_i^j)$ as follows:
$$
h(o) = \left\{ \begin{aligned}
    &1, &&\text{if the object}~o~\text{is hallucinated}; \\
    &0, &&\text{if the object}~o~\text{is not hallucinated}.
\end{aligned} \right.
$$

Based on this evaluation, we categorize the objects into two groups: the hallucinated group $H_1=\{o | o \in O, h(o)=1 \}$ and the non-hallucinated group $H_0=\{o | o \in O, h(o)=0 \}$. We then calculate the fine-grained CLIP score between each object $o_i^j$ in either group and its corresponding image $v_i$. Given that most objects cover only a portion of the image, we segment the image into patches and employ a sliding window technique to identify the best match. Thus, the image-relevance score for each object is determined as follows:
\begin{equation}
    S(o_i^j) = \max_{w_i \in W_i}{\text{CLIP}(o_i^j, w_i)},
\end{equation}
where $W_i$ represents the set of sliding windows over the patches of the image $v_i$.

This methodology enables us to obtain two sets of image-relevance scores $S_1=\{S(o)|o\in H_1\}$ and $S_0=\{S(o)|o\in H_0\}$. In the next section, we will examine the distributions of these scores and validate our hypothesis that text-image similarity can indicate the likelihood of hallucination.

\subsection{Results and Analysis}
\begin{table}[t]
    \centering
    \scalebox{0.95}{
    \begin{tabular}{lcccc}
        \toprule
        Model & Hal. & Mean & Std. & p \\
        \midrule
        \multirow{2}*{MiniGPT4} & No  & 28.26 & 2.74 & \multirow{2}*{$6.0\times 10^{-30}$} \\
                                & Yes & 25.35 & 2.70                    \\
        \midrule
        \multirow{2}*{LLaVA}    & No  & 28.64 & 2.65 & \multirow{2}*{$2.5\times 10^{-12}$} \\
                                & Yes & 26.11 & 2.27                    \\
        \bottomrule
    \end{tabular}}
    \caption{Statistics and significance test on samples generated by MiniGPT4 and LLaVA. Hal. indicates whether the objects are hallucinated, Mean and Std. represent their average and standard deviation of image-relevance scores, and p is the p-value of t-test.}
    \label{tab:pre}
\end{table}

\label{pre_result}
In our analysis, we applied a two-sample t-test to examine the differences between the score distributions of hallucinated and non-hallucinated objects. The results, as detailed in Table \ref{tab:pre}, reveal a notable discrepancy between the mean values of these distributions, as indicated by the p-value. This statistical evidence allows us to confidently reject the null hypothesis that the two distributions have identical means, underscoring the utility of CLIP similarity scores in detecting hallucinations.

To provide a clearer understanding of these differences, we visualized the score distributions through density plots. These plots, illustrated in Figure \ref{fig:pre}, demonstrate that scores for hallucinated objects typically fall below 32, whereas scores for non-hallucinated objects generally exceed 23 for both the two models. Our quantitative analysis further reveals that among the objects scoring above 32, only 0.6\% and 1.6\% are hallucinated, and among those below 23, only 2.3\% and 1.7\% are not hallucinated, for MiniGPT and LLaVA respectively. These findings not only substantiate our hypothesis but also suggest that definitive thresholds can be established to effectively segregate positive and negative samples for the purpose of unlearning.

\section{Multimodal Hallucination Mitigation}
\begin{figure*}[t]
    \centering
    \includegraphics[width=\textwidth]{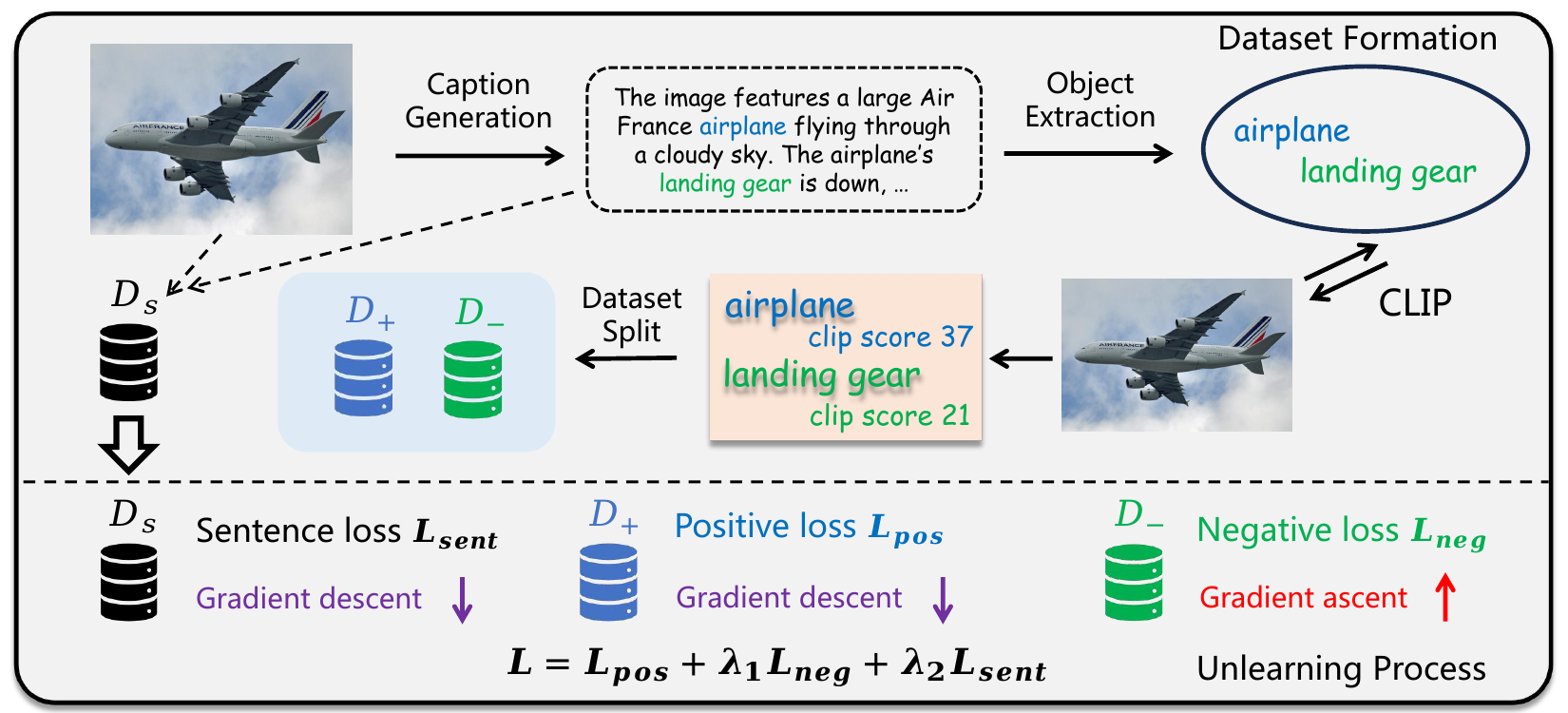}
    \caption{An overview of EFUF. EFUF is divided into two stages: dataset formation and unlearning process. Initially, we extract objects from generated captions and calculate their image relevance utilizing CLIP, followed by the construction of three datasets. Subsequently, three corresponding losses are tailored to finetune the model.}
    \label{fig:method}
\end{figure*}

\subsection{Overview}
After ascertaining the capability of CLIP through a preliminary experiment, we design EFUF, whose overview is shown in Figure \ref{fig:method}. Drawing from established methodologies in prior research \citep{1,2,3,4,5}, our approach is bifurcated into two key stages: dataset construction and the unlearning process itself. Initially, we harness CLIP scores to identify and segregate various samples; after that, unlearning is applied on the model with the curated samples.

Concretely, in constructing the dataset, we first prompt the model to generate captions for given images. After that, we utilize the CLIP model to calculate the fine-grained similarity score of the object phrases in text and the corresponding segments in image. By setting thresholds for the scores, we are able to discern and compile distinct samples from the generated text, forming a dataset for finetuning that circumvents the need for labor-intensive manual annotation. During the finetuning phase, we employ an efficient unlearning method, which involves the development of three distinct types of losses. These losses are designed to aid the model in discarding incorrect multimodal alignments that could lead to hallucinations, while preserving the correct alignments essential for tasks. Unlearning generally requires less computation resources compared with conventional alignment algorithms in the finetuning stage, so the computation amount can also be effectively reduced.

\subsection{Dataset Formation}
Prior to implementing unlearning with MLLMs, it's imperative to define the targets of unlearning and accordingly assemble the requisite positive and negative samples. As evidenced in Section \ref{pre_result}, specific thresholds can effectively delineate between these samples. Hence, we apply these pre-determined image-relevance thresholds to filter the hallucinated and non-hallucinated objects.

Given that a single response may encompass both hallucinated and non-hallucinated objects, a fine-grained approach to unlearning is warranted. Rather than attempting to unlearn an entire response wholesale, we opt for a targeted strategy focusing on the subsentences corresponding to the object, delineated by punctuation. Moreover, to preserve the model’s overarching sentence comprehension and capabilities, we also compile samples of the complete sentences based on the mean image-relevance scores of all included objects, in addition to the positive and negative subsentences. These three categories of samples collectively form the dataset tailored for the unlearning process, facilitating a more nuanced and effective mitigation of multimodal hallucinations.

Formally, let $D=\{v;x;y\}$ denotes a finetuning dataset for MLLM, where $v$ is the image, $x$ is the text query (prompt), and $y$ is the text answer. The positive subsentence dataset is formulated as
$$
D^+ = \left\{ v_i; \text{pre}(o_i^j); \text{cur}(o_i^j) | o_i^j \in O, S(o_i^j) > T_0 \right\},
$$
where $\text{cur}(o)$ represents the subsentence where object $o$ situates, $\text{pre}(o)$ represents all the texts before $\text{cur}(o)$, including prompt, and $T_0$ is the threshold for positive samples. The text that comes after $\text{cur}(o)$ is truncated and unused. Similarly, The negative subsentence dataset is defined as
$$
D^- = \left\{ v_i; \text{pre}(o_i^j); \text{cur}(o_i^j) | o_i^j \in O, S(o_i^j) < T_1 \right\},
$$
where $T_1$ is the threshold for negative samples.


To construct a comprehensive dataset featuring complete responses, it is essential to establish a metric for assessing sentence-level hallucinations. This is achieved by calculating the average image-relevance score across all referenced objects within a response. The formula for this sentence-level image-relevance score is given by
\begin{equation}
    S(t_i)=\frac{1}{n} \sum_{j=1}^n S(o_i^j).
\end{equation}
With this metric, we can curate a dataset of responses by filtering out those responses from the model that meet the specific criterion:
$$
D^s = \left\{ v_i; p_i; t_i | t_i \in T, S(t_i) > T_2 \right\},
$$
where $p_i$ denotes the prompt for response $t_i$, and $T_2$ is the threshold for response samples.

Finally, we take $D_{unlearning}=\{D^+,D^-,D^s\}$ as our unlearning dataset.

\subsection{Unlearning for MLLM}
\label{sec:unlearn}
After constructing the dataset, the final phase of our approach is the application of unlearning techniques to the model. Prior studies \citep{DBLP:journals/corr/abs-2310-02238} have shown that employing solely the unlearning loss severely undermines the model's linguistic comprehension, rendering it incapable of producing coherent sentences. Thus, we introduce a dual-faceted fine-grained unlearning approach: applying a negative loss to the subsentences containing hallucinated objects, and a positive loss to those containing non-hallucinated objects. This strategy aims to curtail the production of hallucinated content while encouraging precise object representation, thus diminishing the occurrence of hallucinations. Meanwhile, we also propose a sentence loss, aiming to preserve the model's ability to generate cohesive, long-form text. In the following, we will introduce these losses in detail.

As is indicated by previous works, the core of unlearning is the gradient ascent strategy. Formally, unlearning updates the model parameters by:
\begin{equation}
    \label{eq:unlearn}
    \Delta \theta = \eta \nabla_\theta L_{ft}(v,x,y;\theta),~~~~~~(v,x,y)\sim D,
\end{equation}
where $\theta$ denotes the model's parameters, $\eta$ is the (un)learning rate, and $L_{ft}$ signifies the finetuning loss function. In the context of multimodal large language models, the supervised finetuning loss function $L$ is articulated as
\begin{equation}
    \label{eq:sft}
    L_{ft}(v,x,y;\theta)=\frac{1}{|y|}\sum_{i=1}^{|y|}l(f_\theta(v,x,y_{<i}),y_i),
\end{equation}
where $f_\theta$ symbolizes the model with parameter $\theta$, and $l(\hat{y}_i,y_i)$ calculates the cross-entropy loss for the predicted and actual values.

To counteract hallucinations while maintaining overall model efficacy, we introduce three distinct losses tailored to the datasets we've constructed. The first, termed \emph{negative loss}, applies gradient ascent to negative subsentences as follows:
\begin{equation}
    L_{neg}=\textcolor{red}{-}L_{ft}(v,x,y),~~~~~~(v,x,y)\sim D^-.
\end{equation}
This inversion of the loss function enables gradient ascent. The second, the \emph{positive loss}, aims at encouraging the model to generate correct objects, with its formulation remaining straightforward:
\begin{equation}
    L_{pos}=L_{ft}(v,x,y),~~~~~~(v,x,y)\sim D^+.
\end{equation}

The last, the \emph{sentence loss} is designed to retain model's comprehension and capabilities on full sentences during the unlearning process:
\begin{equation}
    L_{sent}=L_{ft}(v,x,y),~~~~~~(v,x,y)\sim D^s.
\end{equation}
The overall loss equation then becomes a weighted amalgamation of these three components:
\begin{equation}
    L=L_{pos}+\lambda_1 L_{neg}+\lambda_2 L_{sent},
\end{equation}
where $\lambda_1$ and $\lambda_2$ represent the unlearning weight and the sentence weight respectively.

During training, we perform concurrent sampling from the three datasets, individual loss computation, and aggregation to derive the final loss metric. By doing so, we effectively mitigate hallucinations and preserve the model's proficiency in processing extensive sentences.

\begin{table*}[t]
    \centering
    \scalebox{0.84}{
    \begin{tabular}{lccccccccccc}
        \toprule
        \multirow{2}*{Model} & \multicolumn{5}{c}{Hallucination Rate} && \multicolumn{5}{c}{Generation Quality} \\
        & Chair$_S$$\downarrow$ & Chair$_I$$\downarrow$ & Human$_S$$\downarrow$ & Human$_I$$\downarrow$ & POPE$\uparrow$ && Bleu1$\uparrow$ & Bleu2$\uparrow$ & Bleu4$\uparrow$ & Info.$\uparrow$ & ppl.$\downarrow$ \\
        \midrule
        MiniGPT4
        & 45.9 & 23.2 & 69.0 & 27.3 & 81.0 && 43.8 & 29.5 & 15.5 & 86.7 & 0.134 \\
        \quad + \emph{EFUF}      
        & 38.9 & 21.1 & 45.0 & 12.7 & 82.3 && 45.6 & 31.1 & 16.7 & 87.5 & 0.121 \\
        \midrule
        LLaVA     
        & 52.8 & 22.8 & 42.0 & 14.7 & 85.3 && 43.2 & 29.0 & 15.2 & 93.7 & 0.139 \\
        \quad + \emph{EFUF}      
        & 41.9 & 18.7 & 24.0 & 7.7  & 85.9 && 45.3 & 31.0 & 16.8 & 93.5 & 0.129 \\
        \midrule
        mPLUG-owl
        & 71.1 & 33.5 & 60.0 & 24.1 & 88.5 && 43.3 & 29.1 & 15.1 & 91.1 & 0.129 \\
        \quad + \emph{EFUF}      
        & 40.5 & 23.2 & 46.0 & 17.7 & 90.7 && 52.3 & 35.3 & 19.9 & 90.0 & 0.139 \\
        \midrule
        ShareGPT4V
        & 46.8 & 22.3 & 31.0 & 9.9 & 87.8  && 43.3 & 29.2 & 15.4 & 89.6 & 0.157 \\
        \quad + \emph{EFUF}      
        & 36.9 & 18.4 & 14.0 & 5.4 & 88.1  && 46.9 & 32.5 & 18.1 & 91.1 & 0.159 \\
        \bottomrule
    \end{tabular}}
    \caption{Performance comparison of various MLLMs with and without EFUF. Hallucination is assessed using CHAIR (Chair$_S$, Chair$_I$), MHumanEval (Human$_S$, Human$_I$), and POPE metrics. Quality is evaluated based on consistency with ground truth (Bleu1, Bleu2), informativeness (Info.), and fluency (ppl.). A downward arrow ($\downarrow$) indicates that lower values are better, whereas an upward arrow ($\uparrow$) signifies that higher values are preferable.}
    \label{tab:main}
\end{table*}

\section{Experiments}


\subsection{Experimental Settings}

\paragraph{Dataset.}
We adopt MSCOCO \citep{DBLP:conf/eccv/LinMBHPRDZ14} as our dataset. Since our approach necessitates only the images themselves, their annotations are used exclusively for evaluation. Details of our dataset can be found in Appendix \ref{app:data}.


\paragraph{Evaluation Metrics.}
Following \citet{2}, our assessment encompasses two dimensions: trustworthiness measured by the degree of hallucination, and helpfulness determined by the quality of the generated text. To quantify hallucinations, we utilize CHAIR \citep{DBLP:conf/emnlp/RohrbachHBDS18}, MHumanEval \citep{2} and POPE \citep{DBLP:journals/corr/abs-2306-13394}. For generation quality, we leverage the BLEU \citep{bleu} score for assessing the consistency with ground truth, evaluate informativeness through GPT-4's judgment \citep{gpt4}, and use GPT-2's perplexity score \citep{gpt2} to determine text fluency. Details on the evaluation metrics are provided in Appendix \ref{app:metric}.

\subsection{Baselines}
To affirm the robustness of EFUF across a spectrum of MLLMs, we conducted evaluations against a suite of state-of-the-art base models. These include MiniGPT4 \citep{minigpt}, mPLUG-owl \citep{owl}, LLaVA \citep{llava}, and ShareGPT4V \citep{share4v}, which are pre-trained on extensive multimodal datasets and subsequently finetuned on high-quality instructions. In our experiments, we integrate EFUF into them to obtain the enhanced model.

\section{Results and Analysis}

\begin{table*}[t]
    \centering
    \scalebox{0.81}{
    \begin{tabular}{lccccccccccc}
        \toprule
        \multirow{2}*{Method} & \multicolumn{5}{c}{Hallucination Rate} && \multicolumn{5}{c}{Generation Quality} \\
        & Chair$_S$$\downarrow$ & Chair$_I$$\downarrow$ & Human$_S$$\downarrow$ & Human$_I$$\downarrow$ & POPE$\uparrow$ && Bleu1$\uparrow$ & Bleu2$\uparrow$ & Bleu4$\uparrow$ & Info.$\uparrow$ & ppl.$\downarrow$ \\
        \midrule
        MiniGPT4                                
        & 45.9 & 23.2 & 69.0 & 27.3 & 81.0 && 43.8 & 29.5 & 15.5 & 86.7 & 0.134 \\
        \quad + \emph{unlearn.}                 
        & 42.4 & 22.7 & 56.0 & 17.3 & 82.0 && 44.2 & 29.8 & 15.6 & 87.6 & 0.120 \\
        \quad + \emph{f.g. unlearn.}
        & 36.1 & 17.9 & 39.0 & 9.7  & 82.7 && 47.3 & 32.8 & 17.1 & 87.2 & 0.170 \\
        \quad + \emph{sentence loss}            
        & 44.1 & 29.8 & 58.0 & 17.0 & 81.7 && 43.6 & 29.1 & 16.0 & 86.8 & 0.120 \\
        \quad + \emph{EFUF}                     
        & 38.9 & 21.1 & 45.0 & 12.7 & 82.3 && 45.6 & 31.1 & 16.7 & 87.5 & 0.121 \\
        \bottomrule
    \end{tabular}}
    \caption{Performance comparison of EFUF with vanilla unlearning strategy (\emph{unlearn.}), fine-grained unlearning strategy (\emph{f.g. unlearn.}), and sentence-loss-only method (\%). Although fine-grained unlearning achieves the lowest hallucination rate, it drastically sacrifices fluency, making the generated content difficult for humans to read.}
    \label{tab:ablation}
\end{table*}

\begin{table*}[t]
    \centering
    \scalebox{0.81}{
    \begin{tabular}{lccccccccccc}
        \toprule
        \multirow{2}*{Method} & \multicolumn{5}{c}{Hallucination Rate} && \multicolumn{5}{c}{Generation Quality} \\
        & Chair$_S$$\downarrow$ & Chair$_I$$\downarrow$ & Human$_S$$\downarrow$ & Human$_I$$\downarrow$ & POPE$\uparrow$ && Bleu1$\uparrow$ & Bleu2$\uparrow$ & Bleu4$\uparrow$ & Info.$\uparrow$ & ppl.$\downarrow$ \\
        \midrule
        LLaVA     
        & 52.8 & 22.8 & 42.0 & 14.7 & 85.3 && 43.2 & 29.0 & 15.2 & \textbf{93.7} & 0.139 \\
        \quad + \emph{RLHF}
        & 60.2 & 24.8 & 40.0 & 12.7 & \textbf{87.0} && 39.8 & 25.8 & 12.6 & \uline{93.5} & \textbf{0.126} \\
        \quad + \emph{HADPO}
        & 52.3 & 21.6 & \uline{28.0} & 10.8 & 84.2 && 43.8 & 29.6 & \uline{15.7} & 91.4 & 0.148 \\
        \quad + \emph{POVID}
        & \textbf{41.3} & \uline{19.2} & 29.0 & \uline{8.3} & \uline{86.3} && \uline{44.5} & \uline{30.0} & 15.1 & 86.8 & 0.233 \\
        \quad + \emph{EFUF}      
        & \uline{41.9} & \textbf{18.7} & \textbf{24.0} & \textbf{7.7}  & 85.9 && \textbf{45.3} & \textbf{31.0} & \textbf{16.8} & \uline{93.5} & \uline{0.129} \\
        \bottomrule
    \end{tabular}}
    \caption{Performance comparison of different hallucination mitigation methods for LLaVA on metrics measuring hallucination rate and generation quality. Best scores are in bold and second bests are underlined.}
    \label{tab:compare_1}
\end{table*}

\begin{table}[t]
    \centering
    \scalebox{0.81}{
    \begin{tabular}{lcccc}
        \toprule
        Method & MME & GQA & SQA & QBench \\
        \midrule
        LLaVA
        & \textbf{1491} & \uline{63.0} & 66.9 & \uline{59.2} \\
        \quad + \emph{RLHF}
        & 1212 & 48.4 & 65.4 & 53.0 \\
        \quad + \emph{HADPO}
        & 1441 & 61.2 & \uline{67.2} & 58.6 \\
        \quad + \emph{POVID}
        & 1438 & 61.9 & \textbf{68.4} & \uline{59.2} \\
        \quad + \emph{EFUF}      
        & \uline{1468} & \textbf{63.2} & 66.4 & \textbf{59.3} \\
        \bottomrule
    \end{tabular}}
    \caption{Performance comparison of different hallucination mitigation methods for LLaVA on metrics measuring VQA and reasoning capability.}
    \label{tab:compare_2}
\end{table}

\subsection{Main Results}
As is shown in Table \ref{tab:main}, we evaluate EFUF across a variety of MLLMs, assessing both the hallucination rate and generation quality.

\paragraph{Hallucination Rate.}
Based on the results, our approach demonstrates a consistent reduction in hallucination rates across all four MLLMs, with an average improvement of approximately 15\% and 5\% on the Chair$_S$ and Chair$_I$ metric, 18\% and 8\% on the Human$_S$ and Human$_I$ metric, and 1\% on the POPE metric.
These findings validate the effectiveness and adaptability of our method, emphasizing its capacity to notably lower hallucination rates across cutting-edge models.

\paragraph{Generation Quality.}
Table \ref{tab:main} also highlights the improvements of EFUF in generation quality. Results show that our method not only reduces the hallucination rate but also enhances overall generation quality. Specifically, it improves BLEU-1 by 4\%, BLEU-2 by 3\%, BLEU-4 by 2\%, informativeness by 1\%, and fluency by 1\%, across the four models. These enhancements stem from two main factors: the unlearning strategy which promotes accurate object generation, and the sentence loss design which enhances fluency.



\subsection{Ablation Study}
Without loss of generality, we select the MiniGPT4 model for the ablation study to investigate the effects of different modules of our proposed method. As outlined in Section \ref{sec:unlearn}, our approach is fundamentally comprised of two key elements: the sentence loss and the unlearning mechanism, which itself includes the negative loss and the positive loss. In order to quantify the contribution of each component, we contrast EFUF against the following configurations:
(1) vanilla unlearning: a strategy employing the coarse-grained unlearning, leveraging both positive and negative entire sentences identified based on their sentence-level image relevance scores;
(2) fine-grained unlearning: the unlearning strategy applied in EFUF, but without the sentence loss;
(3) sentence-loss-only method: a method that solely applies the sentence loss of EFUF, omitting the unlearning aspects.
The subsequent content details the outcomes and insights derived from these experimental comparisons.

\paragraph{Effects of Unlearning.}
As shown in Table \ref{tab:ablation}, we observe marginal improvements in hallucination rate reduction and BLEU score enhancement, when the method of vanilla unlearning and sentence loss are applied. However, these gains are trivial compared to those achieved by fine-grained unlearning and the complete EFUF, highlighting the essential role fine-grained unlearning plays in mitigating hallucinations and generating correct objects.

\paragraph{Effects of the Sentence Loss.}
Compared to EFUF, the fine-grained unlearning approach results in a slightly lower hallucination rate but at the cost of informativeness and fluency. In this scenario, BLEU scores fall short of capturing this issue, as they only measure n-gram matches. The decline in fluency is highlighted by a significant increase in perplexity, rendering the responses largely unreadable by humans. Manual examination further reveals that the generated content often consists fragmented and incoherent sentences. Conversely, method employing only the sentence loss and EFUF do not exhibit these flaws, emphasizing the vital function of sentence loss in maintaining high-quality text generation.

In summary, our analysis confirms the necessity of integrating both fine-grained unlearning and sentence loss to effectively reduce hallucinations without compromising the model's proficiency in generating comprehensive, fluent sentences. This combined approach ensures model performance while notably reduces hallucinations.

\subsection{Comparison with Other Methods}
\label{app:compare}
To further evaluate the performance of EFUF, we compare it with other methods tailored to hallucination mitigation. These include LLaVA-RLHF \citep{1}, HA-DPO \citep{4}, and POVID \citep{zhou2024aligning}, which are all evaluated using their officially released checkpoints. We benchmark EFUF against these methods on the LLaVA model, since their checkpoints are all based on LLaVA.

\paragraph{Hallucination Rate \& Generation Quality.}
We measure EFUF's generation quality along with hallucination rate in Table \ref{tab:compare_1}. Compared to other hallucination mitigation methods, EFUF demonstrates comparable or superior performance, while requiring minimal data construction cost and training resources among all. Additionally, our improvements in generation quality are on par with RLHF-based methods, which typically demand expensive human annotations and significant computations. These outcomes highlight our method's effectiveness and efficiency.

\paragraph{VQA \& Reasoning Capability.}
To provide a more holistic evaluation of EFUF, we also assessed its performance on VQA and reasoning tasks. We employed benchmarks such as MME \citep{fu2024mme}, GQA \citep{hudson2018gqa}, ScienceQA \citep{lu2022learn}, and QBench \citep{wu2024qbench}. Table \ref{tab:compare_2} reports the results for the baseline model, EFUF, and competing methods. EFUF demonstrates modest performance fluctuation across these benchmarks compared to other hallucination mitigation strategies, indicating that our method does not negatively affect VQA and reasoning capabilities.

\subsection{Training Cost}
\begin{figure}[t]
    \centering
    \includegraphics[width=0.48\textwidth]{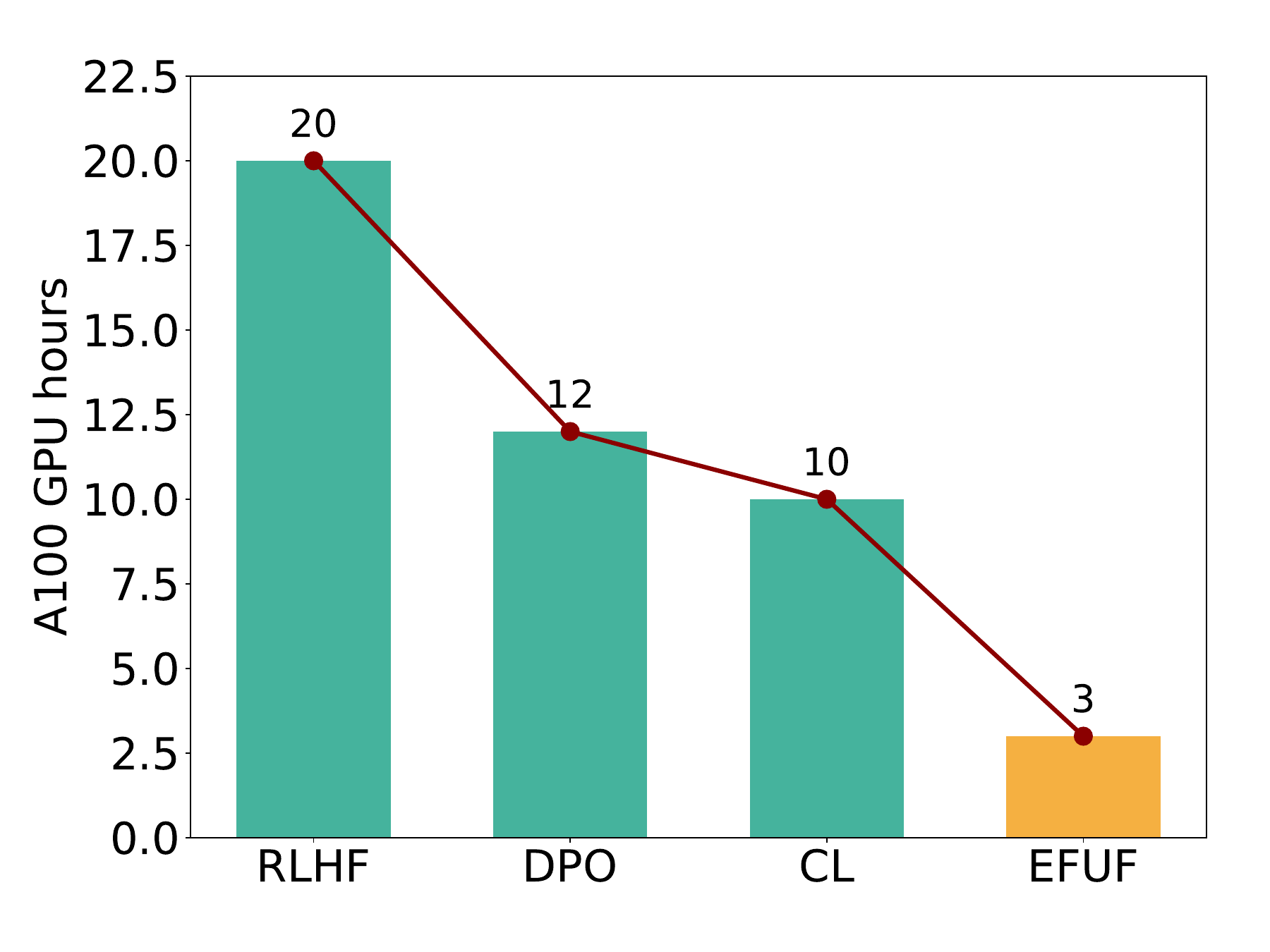}
    \caption{Training time comparison of EFUF with other finetuning-based methods (A100 GPU hours).}
    \label{fig:cost}
\end{figure}

\label{app:cost}
EFUF distinguishes itself from conventional finetuning approaches to hallucination mitigation through its markedly lower end-to-end training costs. A key advantage of EFUF lies in its dataset construction process, which obviates the need for costly human annotations. Traditional methods typically rely on extensive human-labeled datasets, often comprising around 10,000 samples at expenses surpassing \$3,000 \citep{1,2}. Otherwise, they create the dataset with the assistance of GPT-4, involving up to 500,000 samples pre-screened before manual review, incurring costs for around 200 million tokens equivalent to \$2,000 \citep{3,5}.

In stark contrast, EFUF's resource efficiency extends to its training demands. As depicted in Figure \ref{fig:cost}, EFUF's training on an A100 GPU for a MiniGPT4 model requires merely 3 GPU hours, a fraction of the resources needed by other methods. For comparison, RLHF-based finetuning typically consumes 20 GPU hours \citep{1}, DPO ranges from 8 \citep{2} to 16 \citep{4} GPU hours, and contrastive learning method requires around 10 GPU hours \citep{5}.

This substantial reduction on resource requirements in both dataset construction and training stage not only makes EFUF a cost-effective approach but also enhances its scalability and accessibility for broader applications in hallucination mitigation within the realm of multimodal large language models.

\subsection{Additional Analyses}
To further substantiate the effectiveness of EFUF, we provide extensive supplementary analyses in the appendices. As presented in Appendix \ref{app:experiment}, EFUF complements and enhances the performance of existing hallucination mitigation strategies. We also explore the impact of varying weights as hyper-parameters in Appendix \ref{app:parameter}. Finally, a case study detailed in Appendix \ref{app:case} quantitatively evaluates the generated text under different methods, showcasing the distinct advantages of our proposed solution.

\section{Conclusion}
In this paper, we find that text-image similarity is helpful for identifying multimodal hallucinations, and propose a novel unlearning framework to mitigate hallucinations in MLLM. Specifically, we first curate different samples utilizing the image-relevance score derived from CLIP similarity, and then design three distinct losses to perform unlearning on the curated samples. Extensive experiments on different baselines show that our method effectively reduces multimodal hallucinations while retaining the general performance of the model.

\section*{Limitations}
The limitations of our work mainly contain two aspects. Firstly, the exploration of alternative methods for assessing text-image similarity presents an avenue for further research. Our findings affirm the utility of text-image relevance in constructing datasets for the unlearning process, with the relevance scores derived using the CLIP model. Additional methodologies for determining text-image relevance warrant exploration, which may further optimize the construction of unlearning datasets. Secondly, in line with most preceding research, our investigation primarily addresses object hallucinations, gauged by the presence or absence of the depicted object in the corresponding image. The exploration of other varieties of hallucinations, including but not limited to the attributes or positioning of objects within the image, represents a significant area for future work.

\section*{Acknowledgements}
We would like to thank the anonymous reviewers for their constructive comments. This work was supported by the National Natural Science Foundation of China (No. 62206126 and No. 61976114).

\bibliography{custom}

\appendix

\section{Details on Experiment Settings}
\subsection{Implementation Details}
\label{Implementation_Details}
For dataset construction, in order to efficiently obtain the object set $O$, we prompt the LLaMA-2-70b \citep{llama2} model to extract all the objects from the response text. During training, we only tune each model's multimodal mapping layers, i.e., ones that map image feature to text token embedding. We train each model for a fixed 1 epoch with AdamW \citep{adamw} as the optimizer, and report their performance on test set. We implement all the models with the PyTorch framework \citep{pytorch}, and run experiments on an NVIDIA A100 GPU \citep{cuda}. For hyperparameters, we set the weight of unlearning loss $\lambda_1$ to 0.3, the weight of sentence loss $\lambda_2$ to 0.2, the learning rate $\eta$ to 1e-5, weight decay to 0.05. Based on the analysis in Section~\ref{Preliminary_Experiment}, the threshold for normal object $T_0$ and hallucinated object $T_1$ is set to 32 and 23, respectively. Besides, to ensure that the number of the entire sentence samples is similar to that of the positive and negative subsentences, we set the threshold for entire sentence $T_2$ to 27.5.

\subsection{Dataset}
\label{app:data}
MSCOCO \citep{DBLP:conf/eccv/LinMBHPRDZ14} is a comprehensive dataset, encompassing over 300,000 images across more than 80 categories, each meticulously annotated. Our approach, which leverages text image congruence for alignment, necessitates only the images themselves and their associated prompts, omitting any need for annotations. Following \citet{7, 3}, we randomly select 3,200 images with annotation for validation and testing, ensuring no overlap with the training images to maintain the integrity of our experimental conditions.

\subsection{Evaluation Metrics}
\label{app:metric}
\subsubsection{Metrics on Hallucination Rate}
To quantify the rate of hallucinations, we utilize CHAIR \citep{DBLP:conf/emnlp/RohrbachHBDS18} and MHumanEval \citep{2}, which allow us to measure hallucinations at both the sentence and instance levels for model-generated content. Additionally, POPE \citep{DBLP:journals/corr/abs-2306-13394} is incorporated into our evaluation to directly assess the models via VQA. Details of these metrics are given below.

(1) \textbf{CHAIR.} Caption Hallucination Assessment with Image Relevance (CHAIR, \citealp{DBLP:conf/emnlp/RohrbachHBDS18}) is a widely-used metric for evaluating hallucination. It quantifies hallucination by calculating the ratio of non-existent objects referenced in the model's response to the total number of objects mentioned. It features two variations: CHAIR$_S$ for sentence-level and CHAIR$_I$ for instance-level. Both aim to measure object hallucination, albeit from different perspectives:
\begin{align}
    \text{CHAIR}_I&=\frac{|\{\text{hallucinated objects}\}|}{|\{\text{all objects}\}|}, \\
    \text{CHAIR}_S&=\frac{|\{\text{hallucinated responses}\}|}{|\{\text{all responses}\}|},
\end{align}
where hallucinated responses refer to the responses containing at least one hallucinated objects.

(2) \textbf{MHumanEval.} Recognizing the limitations of CHAIR in covering only a set of pre-defined object categories, we also incorporate human judgment into our evaluation. Following \citep{2}, we select a random subset of 100 responses for expert review to identify hallucinated and non-hallucinated objects. Similar to CHAIR, we report hallucination rates at both the object level and the response level, offering a holistic view of the model's accuracy in depicting real-world objects.

(3) \textbf{POPE.} Consistent with prior studies \citep{4, 5}, our evaluation incorporates the Polling-based Object Probing Evaluation (POPE) methodology \citep{DBLP:conf/emnlp/LiDZWZW23}. POPE leverages an automated segmentation tool to delineate objects within images, subsequently querying the model regarding their presence, as well as introducing random non-existent objects. We present the F1 scores, offering insights into the model's image perception capabilities.

\subsubsection{Metrics on Generation Quality}
Our evaluation of the generated content's quality by MLLM hinges on three key metrics: informativeness, consistency with human responses, and fluency. These metrics collectively assess the output's relevance, alignment, and readability.

(1) \textbf{Informativeness.} Inspired by \citep{2}, this metric assesses the extent to which the generated captions encapsulate the primary elements depicted in the image. Utilizing the rich annotations provided by the COCO dataset, we engage GPT-4 \citep{gpt4} to compare the annotated objects, the ground-truth caption, and the model-generated caption, subsequently assigning a coverage score. This process ensures that the evaluation focuses on the caption's ability to highlight significant image details.

(2) \textbf{Consistency to human response.} The fidelity of model-generated content to human-crafted responses is gauged using the BLEU \citep{bleu} score, which measures the linguistic similarity between the machine's output and expert-written ground truth captions. This metric serves as an indicator of how well the model's responses align with human expectations and standards.

(3) \textbf{Fluency.} The smoothness and natural flow of the text produced by the model are evaluated through its perplexity when processed by a pre-trained GPT-2 \citep{gpt2} model. A lower perplexity score signifies higher text fluency, indicating that the generated narrative is coherent and easily comprehensible, mirroring the linguistic quality of the text.

\begin{table*}[t]
    \centering
    \scalebox{0.83}{
    \begin{tabular}{lccccccccccc}
        \toprule
        \multirow{2}*{Models} & \multicolumn{5}{c}{Hallucination Rate} && \multicolumn{5}{c}{Generation Quality} \\
        & Chair$_S$$\downarrow$ & Chair$_I$$\downarrow$ & Human$_S$$\downarrow$ & Human$_I$$\downarrow$ & POPE$\uparrow$ && Bleu1$\uparrow$ & Bleu2$\uparrow$ & Bleu4$\uparrow$ & Info.$\uparrow$ & ppl.$\downarrow$ \\
        \midrule
        LLaVA-RLHF            
        & 60.2 & 24.8 & 40.0 & 12.7 & 87.0 && 39.8 & 25.8 & 12.6 & 93.5 & 0.126 \\
        \quad + \emph{EFUF}   
        & 59.7 & 24.7 & 38.0 & 12.4 & 88.8 && 40.1 & 26.1 & 12.9 & 93.4 & 0.126 \\
        \midrule
        LRV       
        & 39.4 & 19.9 & 46.0 & 16.0 & 85.1 && 51.8 & 36.6 & 20.5 & 88.4 & 0.129 \\
        \quad + \emph{EFUF}   
        & 37.3 & 19.5 & 45.0 & 15.1 & 85.1 && 51.2 & 36.3 & 20.7 & 87.7 & 0.118 \\
        \bottomrule
    \end{tabular}}
    \caption{Performance comparison of EFUF added on other hallucination mitigating approaches (\%).}
    \label{tab:incremental}
\end{table*}

\section{EFUF is beneficial to other hallucination mitigation methods}
\label{app:experiment}
EFUF stands out not only for its effectiveness and efficiency in dataset construction and training but also for its compatibility with existing hallucination mitigation strategies, such as RLHF and instruction tuning. This compatibility suggests that MLLMs already enhanced with such techniques can further benefit from the integration of EFUF, potentially leading to additional performance improvements.

To validate this proposition, we conduct incremental experiments, selecting models enhanced with RLHF (LLaVA-RLHF, \citealp{1}) and instruction tuning (LRV, \citealp{3}) as our new baseline for comparison. These models are then incrementally trained with EFUF. Results, detailed in Table \ref{tab:incremental}, indicate a notable reduction in hallucination rates post-EFUF application, without compromising the quality of the generated text. This outcome underscores EFUF's value as an additive method, capable of augmenting the performance of MLLMs already subjected to advanced hallucination mitigating techniques.

\begin{table*}[t]
    \centering
    \scalebox{0.86}{\begin{tabular}{lcccccccccccc}
        \toprule
        \multicolumn{2}{c}{\multirow{2}*{Parameter}} & \multicolumn{5}{c}{Hallucination Rate} && \multicolumn{5}{c}{Generation Quality} \\
        && Chair$_S$$\downarrow$ & Chair$_I$$\downarrow$ & Human$_S$$\downarrow$ & Human$_I$$\downarrow$ & POPE$\uparrow$ && Bleu1$\uparrow$ & Bleu2$\uparrow$ & Bleu4$\uparrow$ & Info.$\uparrow$ & ppl.$\downarrow$ \\
        \midrule
        \multirow{4}*{$\lambda_1$}
        & 0.1 & 46.3 & 22.1 & 30.0 & 10.2 & 87.7 && 43.2 & 29.2 & 15.4 & 89.5 & 0.155 \\
        & 0.2 & 38.5 & 19.2 & 20.0 & 7.3  & 88.1 && 44.5 & 30.2 & 16.1 & 91.2 & 0.129 \\
        & 0.3 & 36.9 & 18.6 & 18.0 & 5.2  & 88.2 && 47.5 & 33.1 & 18.4 & 90.9 & 0.154 \\
        & 0.4 & 21.0 & 12.5 & 13.0 & 5.9  & 88.0 && 63.5 & 47.0 & 18.1 & 88.5 & 0.243 \\
        \midrule
        \multirow{3}*{$\lambda_2$}
        & 0.1 & 35.7 & 17.7 & 16.0 & 4.3 & 88.4 && 48.6 & 34.1 & 17.9 & 90.6 & 0.187 \\
        & 0.2 & 36.9 & 18.6 & 18.0 & 5.2 & 88.2 && 47.5 & 33.1 & 18.4 & 90.9 & 0.154 \\
        & 0.3 & 39.4 & 19.6 & 30.0 & 7.8 & 87.9 && 45.9 & 31.7 & 16.8 & 91.0 & 0.152 \\
        \bottomrule
    \end{tabular}}
    \caption{Performance of EFUF on the ShareGPT4V model with different negative loss weight $\lambda_1$ and sentence loss weight $\lambda_2$ (validation set).}
    \label{tab:lambda}
\end{table*}

\section{Effects of different weight}
\label{app:parameter}
In this segment, we delve into the effects of varying the weight assigned to the negative loss $\lambda_1$ and sentence loss $\lambda_2$ on the performance outcomes of ShareGPT4V model when trained using our EFUF strategy. The investigation is aimed at understanding how adjustments in these parameters influence both the reduction in hallucination rates and the overall quality of generated content, with results reported on validation set.

(1) Effects of negative loss weight $\lambda_1$
As summarized in Table \ref{tab:lambda}, as $\lambda_1$ is incremented from 0.1 to 0.4, we initially note enhancements in both hallucination reduction and generation quality metrics, up until a value of 0.2. Beyond this threshold and past the value of 0.3, a new trend emerges: while the rate of hallucinations continues to decline, a noticeable degradation in generation quality become apparent. This is particularly evident in the metrics assessing informativeness and fluency, with the most pronounced effects observed once $\lambda_1$ exceeds 0.4. Our case study further reveals the model's diminishing capacity to construct lengthy, informative sentences at the value of 0.4, suggesting an overly aggressive unlearning weight might inadvertently impair the model's foundational knowledge and capabilities.

Given these findings, a value of 0.3 for $\lambda_1$ is identified as the optimal balance point, effectively minimizing hallucinations without compromising the integrity of generation quality.

(2) Effects of sentence loss weight $\lambda_2$
Contrastingly, the impact of $\lambda_2$ generally mirrors the inverse of $\lambda_1$'s effects. A value of 0.1 yields reduced fluency, suggesting that such a low sentence loss weight fails to exert sufficient influence. Conversely, elevating $\lambda_2$ to 0.3 incites an increase in the hallucination rate. This phenomenon can be attributed to an overly dominant sentence loss weight, which biases the model towards learning entire sentence patterns at the expense of neglecting to unlearn hallucinated content. Consequently, a value of 0.2 for $\lambda_2$ is identified as the optimal setting, striking a balance between minimizing hallucinations and maintaining high-quality sentence generation.

\section{Case Study}
\label{app:case}
\begin{figure*}[t]
    \centering
    \includegraphics[width=0.95\textwidth]{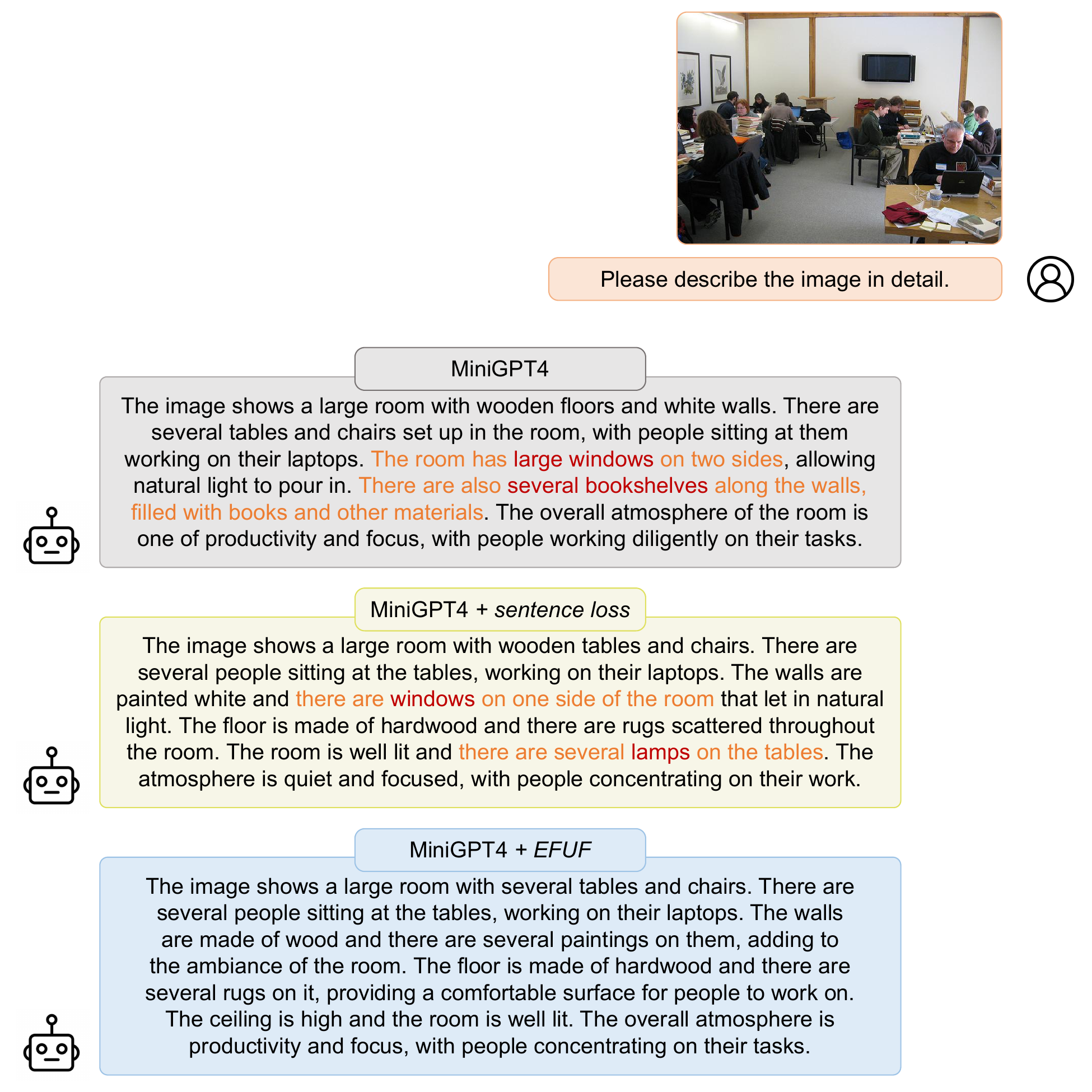}
    \caption{Responses of MiniGPT4 with different methods.}
    \label{fig:case}
\end{figure*}

In this part, we present a comparative analysis through a case study, aiming to elucidate the distinct advantages of our method EFUF. This comparison involves the baseline MiniGPT4 model, a version subjected solely to sentence loss, and the model enhanced with our EFUF strategy.

The case study, as depicted in Figure \ref{fig:case}, highlights a scenario where the base MiniGPT4 model erroneously predicts non-existent elements, such as ``large windows'' and ``bookshelves''. This error is a clear instance of multimodal hallucination, where the generated content includes objects not present in the input image. The sentence-loss-only approach, while attempting to better align the model with multimodal contexts, falls short of completely correcting these hallucinations. This shortfall is attributed to finetuning's inherent limitation: it lacks a mechanism to explicitly signal to the model which objects are inaccurately generated and thus should be excluded from the output.

In contrast, our EFUF approach successfully addresses this challenge. By integrating a fine-grained unlearning strategy, EFUF effectively discourages the generation of objects with low relevance to the given image. This direct intervention ensures that the model refrains from including hallucinated objects in its outputs, showcasing a significant improvement over the baseline and sentence-loss-only method.

\end{document}